\documentclass{article}
\usepackage{iclr2022_conference,times}

\usepackage{amsmath,amsfonts,bm}

\def\eqref#1{equation~\ref{#1}}

\def\1{\bm{1}}

\DeclareMathAlphabet{\mathsfit}{\encodingdefault}{\sfdefault}{m}{sl}
\SetMathAlphabet{\mathsfit}{bold}{\encodingdefault}{\sfdefault}{bx}{n}

\DeclareMathOperator*{\argmin}{arg\,min}

\usepackage{hyperref}
\usepackage{url}

\usepackage{booktabs}
\usepackage{amsfonts}
\usepackage{amsmath}
\usepackage{graphicx}
\usepackage{caption}
\usepackage{nicefrac}
\usepackage{microtype}
\usepackage{xcolor}
\usepackage{bbm}
\usepackage{bm}
\usepackage{enumitem}
\usepackage{gensymb}

\newcommand{\ie}{i.e.,\ }
\newcommand{\eg}{e.g.,\ }

\newcommand{\RN}[1]{\textup{\uppercase\expandafter{\romannumeral#1}}}
\newcommand{\reals}{\mathbb{R}}
\newcommand{\nonneg}{\reals_{+}}
\newcommand{\integers}{\mathbb{Z}}

\newcommand{\mcal}{\mathcal{M}}
\newcommand{\dcal}{\mathcal{D}}
\newcommand{\msf}{\mathsf{M}}

\newcommand{\id}{\mathrm{id}}
\newcommand{\deq}{\overset{d}{=}}

\newcommand{\diffloglen}{\Delta}

\newcommand{\xg}{z}
\newcommand{\alphamat}{{\bm{\alpha}}}
\newcommand{\betamat}{{\bm{\beta}}}
\newcommand{\idmat}{{\bm{\mathrm{id}}}}
\newcommand{\cvxhull}{\mathrm{ConvexHull}}

\title{Learning from One and Only One Shot}
\author{
Haizi Yu$^{1,2}$ \qquad Igor Mineyev$^1$ \qquad Lav R.~Varshney$^1$ \qquad James A.~Evans$^2$
\\
$^1$ University of Illinois Urbana-Champaign \qquad $^2$ University of Chicago
\\
\texttt{\{haiziyu7,mineyev,varshney\}@illinois.edu}, \texttt{{jevans}@uchicago.edu}
}
\iclrfinalcopy

\begin{document}

\maketitle

\begin{abstract}

Humans can generalize from only a few examples and from little pretraining on similar tasks.
Yet, machine learning (ML) typically requires large data to learn or pre-learn to transfer.
Motivated by nativism and artificial general intelligence, we directly model human-innate priors in abstract visual tasks such as character and doodle recognition.
This yields a white-box model that learns general-appearance similarity by mimicking how humans naturally ``distort'' an object at first sight.
Using just nearest-neighbor classification on this cognitively-inspired similarity space, we achieve human-level recognition with only $1$--$10$ examples per class and no pretraining.
This differs from few-shot learning that uses massive pretraining.
In the tiny-data regime of MNIST, EMNIST, Omniglot, and QuickDraw benchmarks, we outperform both modern neural networks and classical ML.
For unsupervised learning, by learning the non-Euclidean, general-appearance similarity space in a $k$-means style, we achieve multifarious visual realizations of abstract concepts by generating human-intuitive archetypes as cluster centroids.

\end{abstract}

\section{Introduction}

Modern machine learning (ML) has made remarkable progress, but this is accompanied by increasing model complexity, with hundreds of neural layers (\eg ResNet-152) and millions to trillions of parameters (\eg ViT: $86$-$632$M, GPT-4: $1$T).
This results in a huge appetite for data and resources, making data curation hard and energy costs irresponsibly high, which particularly challenges domains like low-resource languages or rapidly-evolving pandemics.
The increased model complexity further leads to inscrutability and nonintuitiveness, making the model hard both for users to control and for developers to tune (\eg hyperparameters, architecture).
As such, there is a need for ML models that are \emph{prior-} and \emph{data-efficient}~\citep{Chollet2019}, that are \emph{human-like}~\citep{LakeST2015}, and that exhibit \emph{human-interpretable} behaviors~\citep{AdadiBerrada2018}.

Few-shot learning~\citep{LakeSGT2011,BrownMRSKDNSSA2020,WangYKN2020} via transfer learning~\citep{PanY2009,FinnAL2017,HsuLF2019} has succeeded in some data-scarce scenarios, but requires ``relevance'' between the transferring source and target~\citep{Storkey2009}.
Yet, knowing such relevance in advance and understanding what is transferred are often black arts.
This is especially the case in new, understudied domains, with a risk of unwanted \emph{negative transfers}~\citep{PanY2009,MeiselesR2020}.
There has also been a shift from ``big transfer''~\citep{KolesnikovBZPYGH2020} to ``small transfer''~\citep{LakeST2019}---pressing for reduced pretraining.
We push such reduction to the limit: to achieve human-level generalization ability and interpretability, with no pretraining at all.

These engineering challenges are intertwined with a scientific puzzle: how do humans learn so much from so little~\citep{May2015}?
From growing children to artificial general intelligence, every new challenge involves data-sparsity.
We think from a nativist perspective.
Given a single example, humans \emph{abstract} it and can conceive of further equivalent examples.
Many abstraction abilities are innate~\citep{SpelkeK2007}.
Babies can tell things apart based on \emph{general appearance}, \ie how things look in general (topologically, geometrically, structurally...), knowing they may be translated, rotated, scaled, or deformed.
\citet{SloutskyKF2007} further showed humans' early induction is mainly based on appearance similarity rather than \emph{kind} information.
Babies can easily see similarity between a written ``1'' and a crutch before knowing what they are.
This view of \emph{categorization after appearance similarity} yields our focus on learning such a similarity after which simple ML techniques such as $k$-NN or $k$-means can be used to maintain interpretability.

\begin{figure}[t]
\begin{center}
\includegraphics[width=0.8\columnwidth]{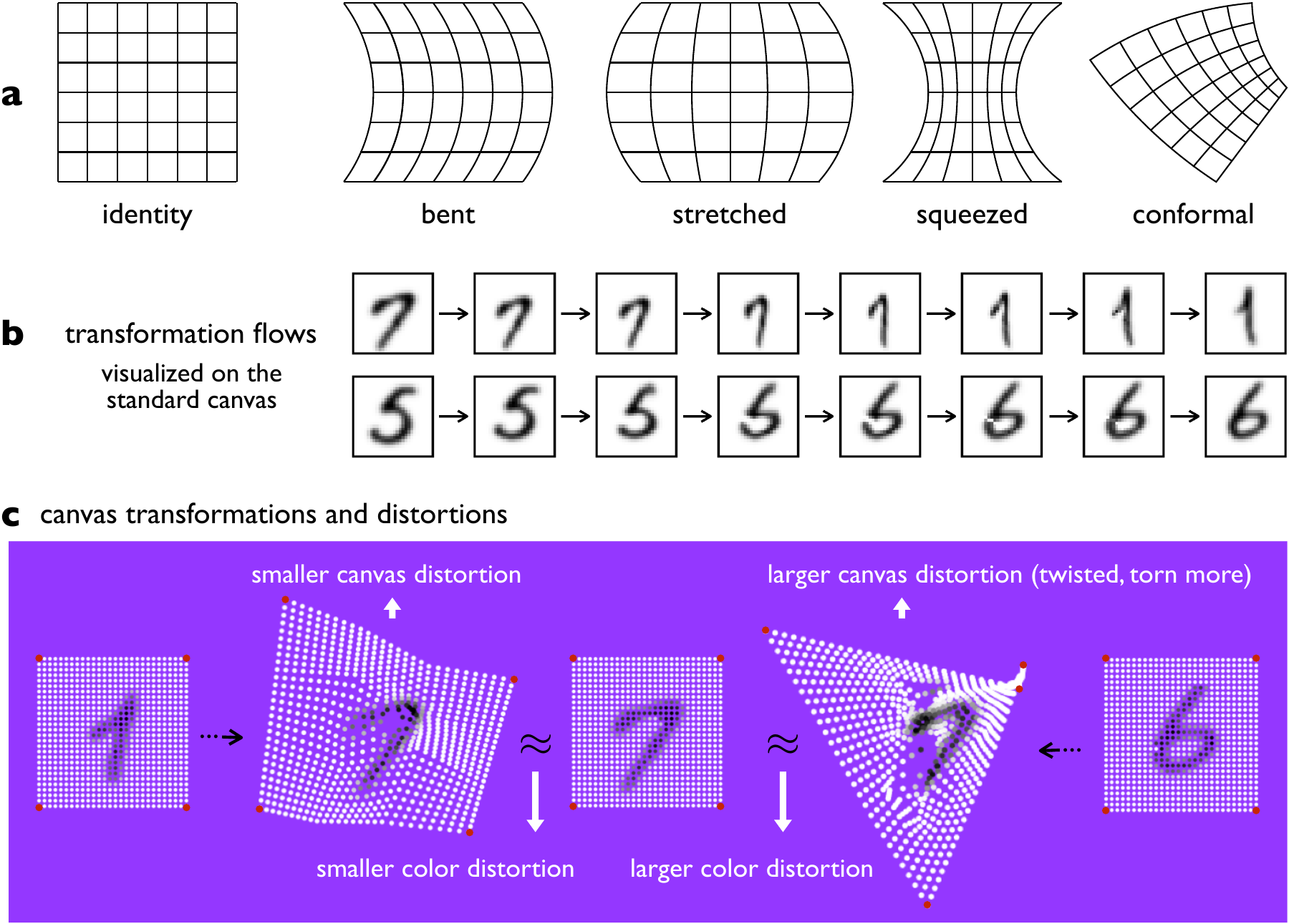}
\end{center}
\caption{Canvas transformations (a), transformation flows (b), and distortions (c).}
\label{fig:transfs-and-flows}
\vspace{-0.05in}
\end{figure}

We present a theoretically sound, white-box model that mimics how humans learn and generalize from only a few examples.
More specifically, we devise a \emph{distortable canvas} to computationally realize the nativist intuition about humans' innate perception.
The idea is to view every image being smoothly painted on an elastic (\eg rubber) canvas that can be distorted in many ways (Figure~\ref{fig:transfs-and-flows}a).
Due to elasticity/viscosity, larger distortions expend more energy, so intuitively, two images have a similar appearance if one can transform into the other with little energy.
This yields our mathematical formulation of \emph{general-appearance similarity} based on minimal \emph{canvas and color distortions}.

We address three main technical challenges in learning this similarity.
First, we parameterize and efficiently handle \emph{all} transformations (including those without a formula) instead of handpicking special ones by domain knowledge (like scale, translation, and rotation invariances commonly used in classical computer vision).
Second, we introduc an \emph{abstracted multi-level gradient descent (AMGD)} method to mimic humans' hierarchical abstraction ability and lift the \emph{curse of local minima} during optimization.
Third, we make gradient descent, and hence the full optimization process, interpretable by eliciting the full transformation flow (Figure~\ref{fig:transfs-and-flows}b) instead of just the final transformation (Figure~\ref{fig:transfs-and-flows}c).
These flows produce insights that either match our intuition (``Yes, that is what I would naturally do to transform 7 into 1.'') or unveil new perspectives (``I did not realize this other way of transforming 7 into 1. Now I see it and it makes sense to me.'').

Our distortable canvas shares similar ideas with other transformation-based models using group theory~\citep{YuMV2021,YuMV2023,YuEV2023}, optimal transport~\citep{Vilani2009}, morphing~\citep{LeeWS1998,ClarkeCTM2006}, invariances~\citep{Lowe1999}, or equivariances~\citep{BronsteinBLSV2017}, but we do not handpick or restrict ourselves to special transformations.
Compared to data augmentation~\citep{KrizhevskySH2012imagenet}, we build transformations into the model rather than into the data (and learning from data afterwards): the former may be viewed as the latter under infinite augmentation and perfect learning.

We achieved success on abstract visual tasks such as character and doodle recognition.
On image classification benchmarks including MNIST~\citep{MNIST}, EMNIST~\citep{EMNIST}, as well as the more challenging Omniglot~\citep{LakeST2015} and QuickDraw~\citep{Quickdraw} datasets, simply running the nearest-neighbor method on our learned similarity space outperformed both contemporary neural networks and classical ML in the tiny-data or single-datum regime.
To name a few highlights: with no pretraining, our model reached $80\%$ MNIST accuracy using \emph{only the first} training image per class (reached $90\%$ using only the first four) and achieved near-human performance on Omniglot and QuickDraw one-shot learning tasks.
In unsupervised learning, integrating simply $k$-means into our model captured human-level visual abstractions, which generated human-intuitive archetypes as cluster centroids (\eg different ways of writing ``7'' or doodling a giraffe).

\section{General-Appearance Metric Learning via Distortable Canvas}

Our \emph{distortable canvas} models all images with color (grayscale for now) smoothly painted on an elastic canvas.
Any such smooth image $\mcal: \reals^2 \to \nonneg$ can be flexibly distorted via all kinds of canvas transformations $\alpha: \reals^2 \to \reals^2$ (\eg bent, stretch) and a color transformation $\chi: \nonneg \to \nonneg$, resulting in a freely transformed image $\chi \circ \mcal \circ \alpha$.
We further introduce {\color{black} \emph{canvas distortion}}~$\dcal_V(\alpha)$ for any canvas transformation $\alpha$ and {\color{black} \emph{color distortion}} $\dcal_C(\mcal,\mcal')$ between two smooth images $\mcal,\mcal'$.
Our idea is to search for a transformation that mimics what humans naturally do to transform one image into another.
That is, a low-distortion $\alpha$ that yields little difference in color between $\mcal$ and a transformed $\mcal'$.
More precisely, we want to minimizes both $\dcal_V(\alpha)$ and $\dcal_C(\mcal,\ \chi \circ \mcal' \circ \alpha)$.
This yields two dual variants of our desired general-appearance distance:
\begin{itemize}[leftmargin=0.25in]
\item $\dcal_C$-distance: minimizes the color distortion $\dcal_C$ while controlling the canvas distortion $\dcal_V$
\item $\dcal_V$-distance: minimizes the canvas distortion $\dcal_V$ while controlling the color distortion $\dcal_C$.
\end{itemize}

We relegate the detailed formulations about image smoothing, canvas and color transformations, as well as canvas and color distortion minimizations to Appendix~\ref{sec:smooth-image-on-distortable-canvas}.

Besides a final transformation and distance, we apply the minimal-distortion principle to the entire transformation process, \ie to keep canvas and color distortions small along the entire optimization process~\citep{MesaTMKC2019}.
This yields a smooth transformation flow that mimics human intuition too: our mind does not treat translations as sudden displacements from one location to another, but tends to auto-complete a translation path that is continuous and desirably short.

Gradient descent naturally fits this goal by always pursuing the steepest descent.
However, it suffers from the \emph{curse of local minima}.
Our solution is to lift gradient descent to multiple levels of abstraction via multiscale canvas lattices and color blurring, mimicking human abstraction capabilities that are extremely flexible in multiscale optimization.
We name this technique the \emph{abstracted multi-level gradient descent (AMGD)}, which is controlled by an anchor-grid system $\hat{G}$ and a blurring parameter $\rho_c$ in the optimization.
The result of applying AMGD to solve the optimization problem is not just a final solution but a so-called $(\hat{G},\rho_c)$-solution path, which delineates the backbone of a desired transformation flow.
We relegate more details about AMGD to Appendix~\ref{sec:amgd}.

\section{Image Classification in the Tiny Data Regime}

Using $\dcal_C$- or $\dcal_V$-distance in the nearest-neighbor method yields our $\dcal_C$- or $\dcal_V$-nearest-neighbor classifier.
The transparency of the distortable canvas and the simplicity of nearest-neighbors makes the whole metric-learning and classification process human intuitive and interpretable.
We demonstrate classification performances on hand-drawn characters/doodles from four benchmarks.
These include the MNIST (digits) and EMNIST (letters) datasets restricted to the tiny-data regime, as well as the Omniglot (scripts) and the QuickDraw (doodles) one-shot learning challenges.

\begin{figure}[t]
\begin{center}
\includegraphics[width=1\columnwidth]{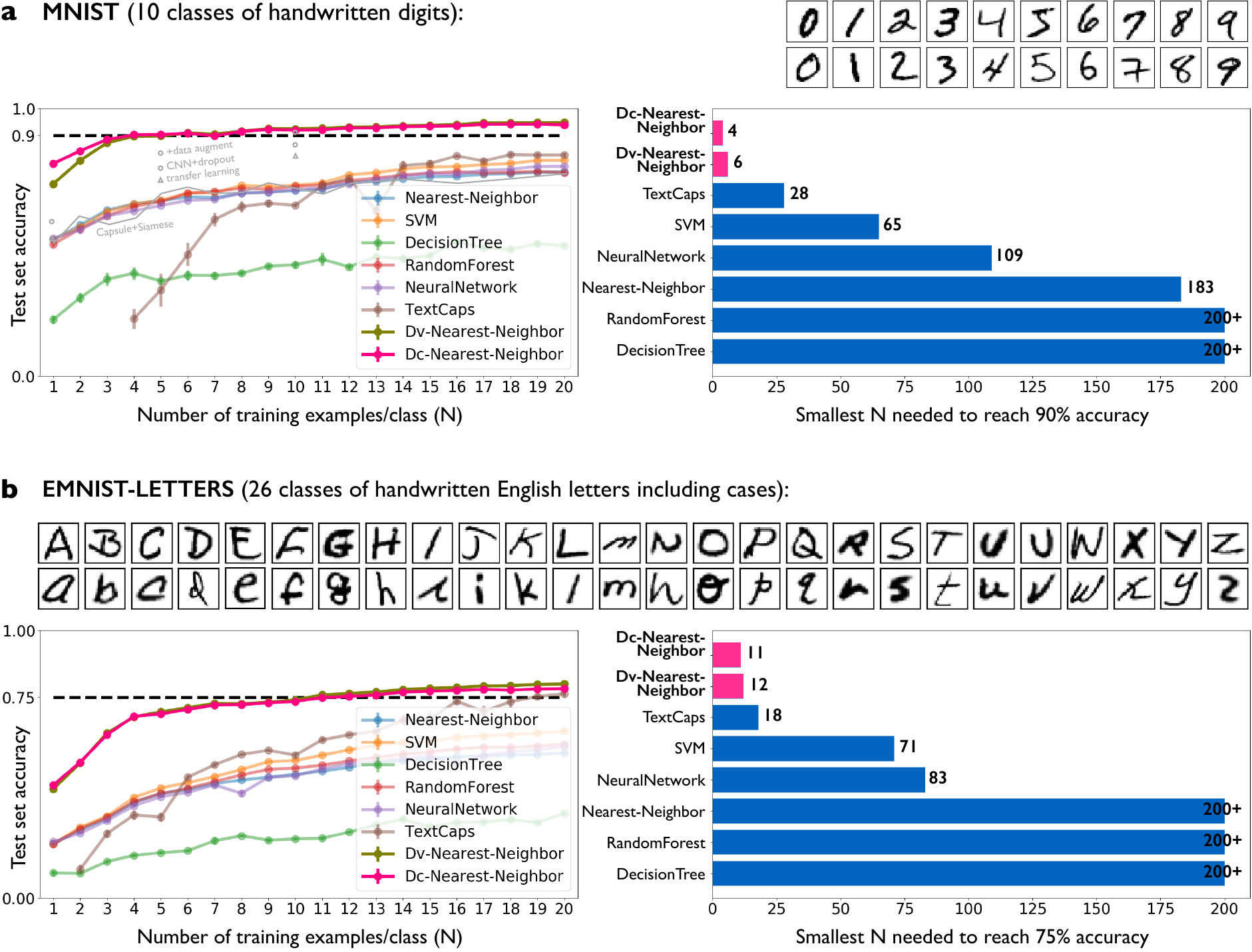}
\end{center}
\caption{MNIST and EMNIST in the tiny-data regime: the first $1$--$20$ training images per class and the full test set are used.
For each classifier, we plot its test accuracy versus the training size $N$ as well as the smallest $N$ needed to reach an accuracy threshold ($90\%$ for MNIST and $75\%$ for EMNIST due to increased difficulty).
Our model outperforms all other comparison models for all $N$, requiring the least amount of training data to perform well.}
\label{fig:mnist-plus-emnist-letters}
\end{figure}

\textbf{MNIST in the tiny-data regime.}
The original benchmark has $60$k images for training and $10$k for testing, spanning $10$ classes.
To evaluate how a model performs in the tiny data regime, we train the model on the first $N$ images per class from the original training set, test it on the full test set, and record test accuracy for $N = 1, 2, 3, \ldots$.
We compare our model to both contemporary neural networks and classical ML models, including TextCaps~\citep{JayasundaraJJRSR2019} with state-of-the-art performance in the small-data regime, SVM, nearest-neighbor, etc.
Classical ML is included to show that success in the tiny-data regime does not mean using simple models.
For stochastic models, we record mean and standard deviation from $5$ independent runs.
TextCaps only runs when $N \geq 4$ and sometimes returns a random guess ($10\%$), so we record trimmed mean and standard deviation from $11$ runs (where we trim the best two and worst four).
We also report results from the literature that ran MNIST in a similar tiny-data setting, including FSL that uses extra data for pretraining (whereas all our other comparison models do not).
These results are from the same training-testing sizes but not the same data sets, and hence are considered indirect comparisons.
We present all results in Figure~\ref{fig:mnist-plus-emnist-letters}a.

\textbf{EMNIST-letters in the tiny-data regime.}
The original benchmark has $4.8$k training images per class and $0.8$k test images per class, spanning $26$ classes of case-insensitive English letters.
We keep the same experimental setting as in MNIST (except for TextCaps being more stable now: we do $7$ independent runs for each $N$ and trim the best and the worst).
Results are shown in Figure~\ref{fig:mnist-plus-emnist-letters}b.
EMNIST-letters is harder, not only with more classes but also more intrinsic ambiguities (\eg \emph{l} and \emph{I}, likewise \emph{h} and \emph{n}, can be written very similarly; while \emph{r} and \emph{R} look different despite their semantic similarity).
So, all models perform significantly worse than in MNIST.
The intrinsic ambiguity, as well as more labeling errors, narrows our superiority over other models as training size increases.
This is especially true for the state-of-the-art TextCaps model, catching up quickly in Figure~\ref{fig:mnist-plus-emnist-letters}b.

Being sensitive to ambiguities and outliers, however, is not a result from our distortable canvas model.
It is due to the nearest-neighbor inference.
To improve, we might consider integrating our model with more robust classifiers, \eg $k$-nearest-neighbor ($k$-NN) with proper voting.
However, $k$-NN is not very effective in the tiny-data regime, not only because the training size can be as small as $k$ but also there is little room to hold out a validation set for selecting $k$.
An adaptive $k$-NN may be desired, with $k$ remaining $1$ in the tiny-data regime and becoming tunable when training size increases to a level that affords a held-out validation set.
A related issue due to lacking validation data is in picking a proper model configuration.
One may expect better results from any comparison model in Figure~\ref{fig:mnist-plus-emnist-letters} by trying new configurations which however can be a black art.
For TextCaps, we used its original implementation and configuration;
for the rest, we used scikit-learn implementations with default configurations (except for tweaks like neural-network size and regularization for the tiny-data setting).
By contrast, our distortable canvas has little to tune, other than the $(\hat{G}, \rho_c)$-solution path in AMGD.
In general, the more gradual the path is, the better.
We picked $(\hat{G}, \rho_c)$ based on runtime and image size ($28\times 28$ here) only.

\begin{figure}[t]
\begin{center}
\includegraphics[width=0.9\columnwidth]{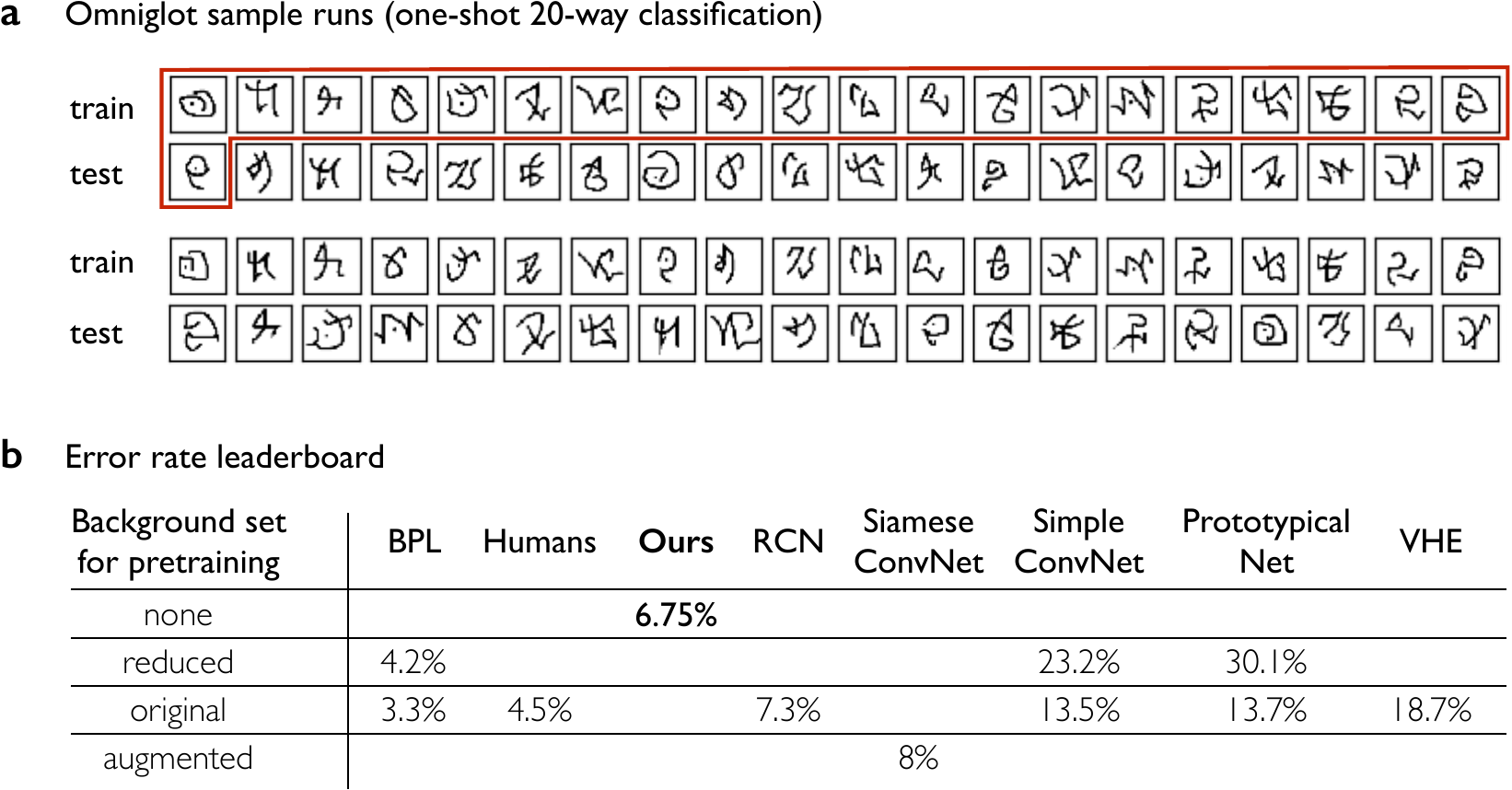}
\end{center}
\caption{Omniglot one-shot classification: two sample runs (a) and the error-rate leaderboard (b). The red outline marks one out of $400$ unit tasks, made up of $1$ test and $20$ training images.}
\label{fig:omniglot}
\end{figure}

\textbf{The Omniglot challenge for one-shot classification}.
The Omniglot dataset contains handwritten characters from $50$ different alphabets, which include historical, present, and artificial scripts (\eg Hebrew, Korean, ``Futurama'') and are far more complex than MNIST digits and EMNIST letters.
The characters are stored as both images and stroke movements.
Unlike MNIST/EMNIST that come with large training data, Omniglot was designed for human-level concept learning from small data.
Its one-shot classification task was benchmarked to evaluate how humans and machines can learn from a single example.
This benchmark contains $20$ independent runs of $20$-way within-alphabet classifications.
The $(2k-1)$th and $(2k)$th runs for $k = 1, \ldots, 10$ use the same set of $20$ characters from a single alphabet.
Each run uses $40$ images: one training and one test image per character.
The unit task here is for each test image, to predict the character class it belongs to (one of $20$), based on the $20$ training images.
In total, there are $400$ independent unit tasks across all $20$ runs.
Figure~\ref{fig:omniglot}a shows a unit task (in red) and the first two runs in the benchmark, covering $1$ alphabet, $20$ characters, and $80$ distinct images.

The Omniglot benchmark adopts the standard FSL setting, where a background set is also provided for pretraining.
The original background set contained $964$ character classes from $30$ alphabets; later, a reduced background set was proposed to make the classification task more challenging.
We dispense with any background set and any stroke information when running our $\dcal_C$-nearest-neighbor.
In each unit task, we predict the test image based on \emph{one and only that} training image per character, and we read all images from raw pixels.
Shown in Figure~\ref{fig:omniglot}b, our model (with a $6.75\%$ error rate) approaches human performance ($4.5\%$) and outperforms all models in the Omniglot leaderboard~\citep{LakeST2019}, except for BPL that was specially designed for the Omniglot challenge by making additional use of both the background set and the stroke information.

\begin{figure}[t]
\begin{center}
\includegraphics[width=0.97\columnwidth]{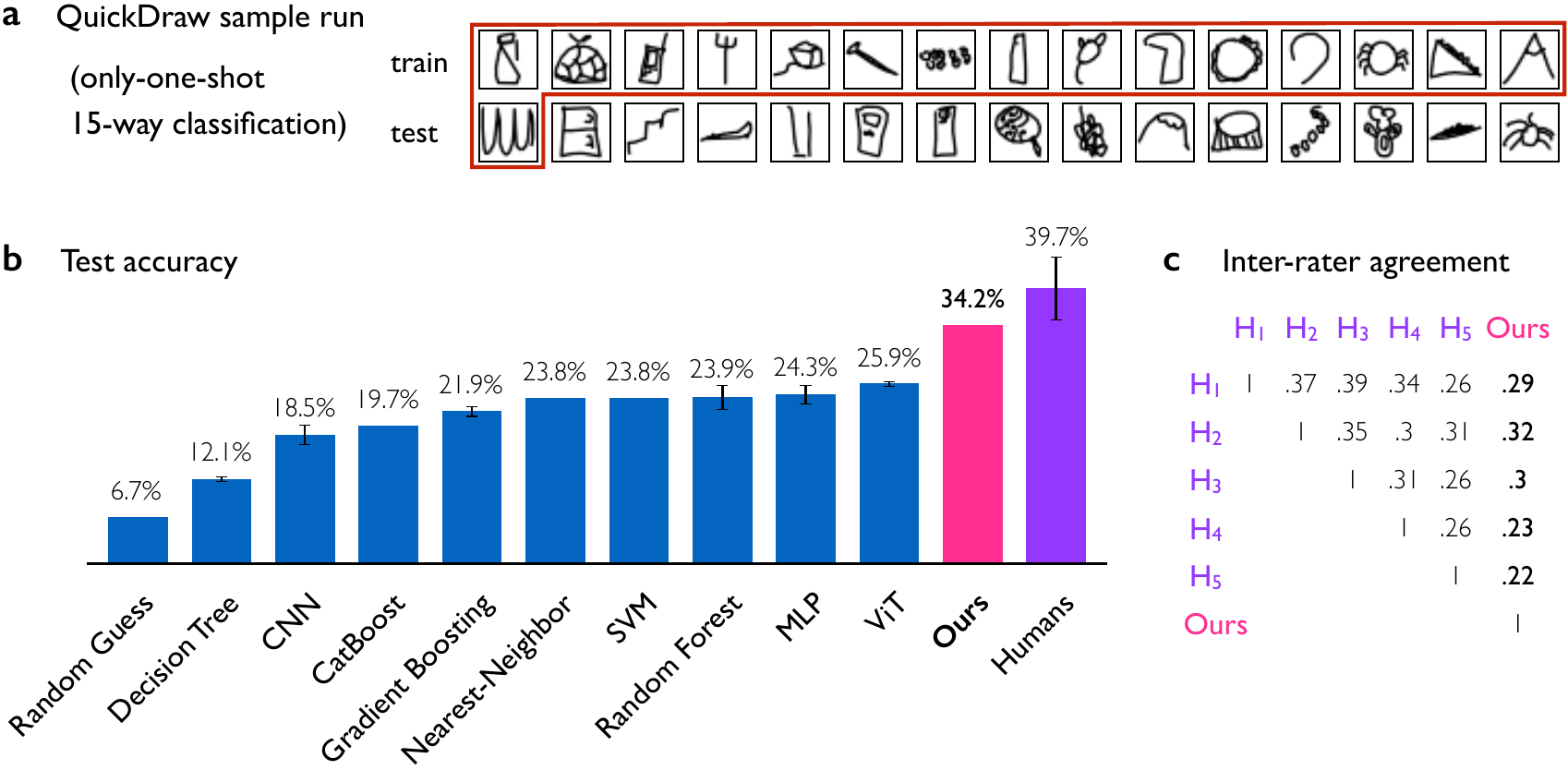}
\end{center}
\caption{QuickDraw only-one-shot doodle classification: one sample run (a), test accuracies (b), and inter-rater agreement between our model and human performances (c). The sample run exemplifies the training and test images, with the red outline marking a unit task. Test accuracy uses the percentage and the error bar to show the mean and the standard deviation from $5$ independent runs of each model (error bars are omitted for deterministic models). The inter-rater agreement shows the pairwise Fleiss' kappa. Human results are from $5$ healthy subjects ($H_1, \ldots, H_5$) between the ages of $20$ and $29$.}
\label{fig:qd1shot}
\end{figure}

\textbf{Human doodles for only-one-shot classification}.
Beyond handwritten characters, we experiment with recognizing human doodles in the only-one-shot setting.
Unlike writing systems designed for people to follow certain ways of writing, there is no ``correct'' way of doodling a particular object or concept in mind---everyone has their own picture of Hamlet and hurricanes.
Further, unlike photos, doodles are often unfaithful visual reproductions of an object's outlook: many doodles only capture core features abstractly.
These attributes make doodle recognition a fundamentally much harder task, even for humans.
In this experiment, we use Google's QuickDraw dataset containing $345$ categories of human doodles.
QuickDraw data are visually much more abstract and difficult than other datasets of human sketches such as Sketchy~\citep{DeyRDLS2019,ChowdhuryBSKXS2023}.

Mimicking the Omniglot setup, we randomly divide all categories into $23$ runs ($15$ categories per run).
In each run, we randomly sample two doodle images from each category---one for training and one for testing---forming a training and a test set, each containing precisely one image per category.
The unit task here is for each test image, to predict the doodle category it belongs to (one of $15$), based on just the $15$ training images.
Every unit task is an independent one-shot $15$-way classification problem, meaning the fact that every test image is from a distinct category is not leveraged.
In total, there are $345$ unit tasks across all runs.
Figure~\ref{fig:qd1shot}a illustrates the training and test images in one run and a unit task (in red).
All models in this experiment read images directly from raw pixels, without using stroke-movement information.
Unlike Omniglot adopting the standard FSL setting, all models here have no access to any pretraining set and hence reside in the only-one-shot setting.

Shown in Figure~\ref{fig:qd1shot}b, our model (with a $34.2\%$ accuracy) approaches human performance ($39.7\%\pm4.6\%$) and outperforms all other models.
Due to the only-one-shot setting, there is no extra room for a validation split normally used for configuration/hyperparameter tuning.
All comparison models in Figure~\ref{fig:qd1shot}b use their standard implementation in ${\tt sklearn}$ or $\tt keras$, \eg the convolution neural network (CNN) and the vision transformer (ViT) are from $\tt keras$' code examples with proven test performance in the standard MNIST and CIFAR-100 benchmark.
Our model is the same $\dcal_C$-nearest-neighbor used in the previous MNIST/EMNIST experiments.
In Figure~\ref{fig:qd1shot}c, we see fair agreement (quantified by Fleiss' kappa in $0.21$--$0.4$) among humans as well as a similar level of agreement between our model and each human.
This indicates that our model performs not just at an accuracy level near humans, but also has fair agreement with humans on the mistakes it makes.

Notably, unlike computational models, human participants in this experiment are not really in the only-one-shot setting.
Unlike babies faced with the doodles for the first time, participants might have unconsciously (or inevitably) used extra knowledge even though they were instructed to try not to.
For example, when a subject looked at the last test image in Figure~\ref{fig:qd1shot}a, (s)he might have first inferred that ``this is a crab or spider'' and then attempted to find a ``crab or spider'' among the training images.
Knowledge of concepts like that of ``a spider'' puts experienced humans at an advantage in this experiment.

\section{Unsupervised Learning: Archetype Generation}

Beyond classification, our distortable canvas enables $k$-means-style clustering in a general-appearance similarity space that is non-Euclidean and human-intuitive.
As in other non-Euclidean metric learning settings~\citep{CuturiD2014}, it is unrealistic to run $k$-means on explicitly computed distances.
Learning a distance in our model requires solving an optimization problem, which is much more expensive than computing Euclidean distances.
Further, computing a centroid in a non-Euclidean space requires solving another optimization problem (minimizing the sum of within-cluster distances), which is much more expensive than an arithmetic mean.
What is more challenging is that the two optimizations are nested, yielding an optimization problem of optimization problems.

To address these challenges, we generalize our idea of a transformation flow between two images into multi-flows among multiple images.
Under this generalization, we do not explicitly compute pairwise distances, meaning we do not solve the inner optimizations first.
Instead, we solve the inner and outer optimizations at the same time, where we flatten the nested optimizations into a single one.
We relegate more details to Appendix~\ref{sec:transformation-multi-flows}.

\begin{figure}[t]
\begin{center}
\includegraphics[width=0.95\columnwidth]{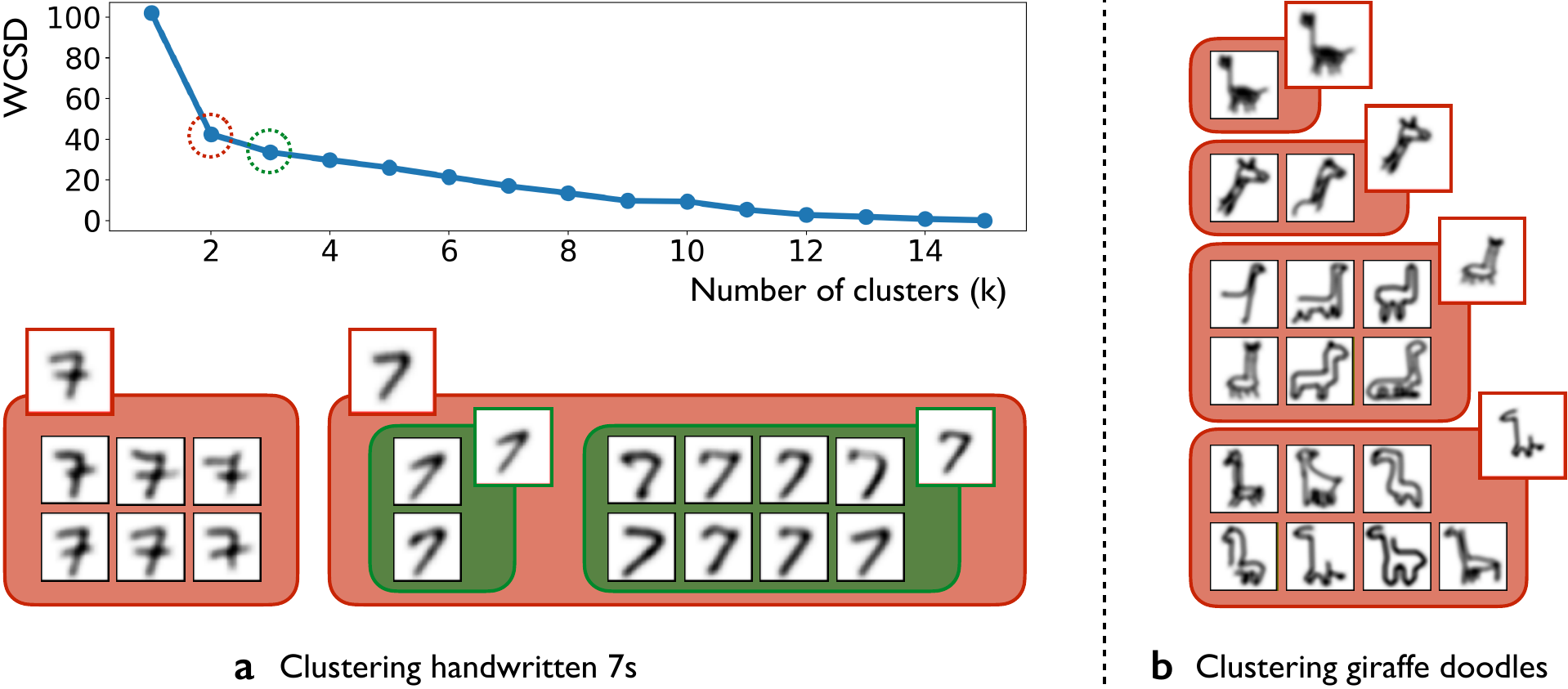}
\end{center}
\caption{Archetype generation via $k$-means-style clustering in our learned similarity space.}
\label{fig:archetypes}
\end{figure}

As in $k$-means, we try different values of $k$, and for each $k$, we try multiple random starts and record the best within-cluster sum of distances (WCSD).
We use the elbow method to pick good $k$-values.
Figure~\ref{fig:archetypes}a shows the WCSD-versus-$k$ curve obtained by running our clustering method on a set of $16$ images of ``7''s from MNIST.
The curve indicates $k = 2$ or $3$ as a potential elbow point.
The resulting two clusters of ``7'' agree with human intuition regarding two general ways of writing ``7'', depending on whether there is an extra stroke.
The resulting three clusters further divide the cluster of ``simpler 7s'' based on the angle of the transverse stroke.
Figure~\ref{fig:archetypes}b shows four clusters of giraffe doodles and their centroids learned from the first $16$ giraffes in Google's QuickDraw.
We see a clear separation of outline sketches, focused views of the neck, and two different pose orientations.

Both examples in Figure~\ref{fig:archetypes} show that the cluster centroids learned from our model can be effectively viewed as \emph{archetypes} of the input images (\eg different ways of writing ``7'' or doodling a giraffe).
These human-intuitive archetype generations demonstrate our model's ability in effective visual abstractions (a strong contender in Pictionary) and further imply their value in education.

\section{Conclusion, Limitation, and Future Work}

This paper designs a white-box model to learn from few and only those few examples---in particular one and only one example---requiring no extra data for pretraining.
Based on nativism, our distortable canvas effectively models human intuition about general appearance and learns transformation-based similarity akin to how humans naturally ``distort'' objects for comparison.
This notion of similarity is formalized in our proposed optimization problem, which minimizes canvas and color distortions to transform one object into another with minimal distortion.
To remedy vanishing gradients and solve the optimization efficiently, we mimic human abstraction ability by chaining anchor lattices and image blurs into a solution path.
This yields our AMGD method capable of optimizing at multiple levels of abstraction.
Our model outputs not only transformations but also transformation flows that mimic efficient human thought processes.
We demonstrate success in benchmarks focused on abstract visual tasks such as character and doodle recognition.
By simply using $1$-NN, we achieve state-of-the-art results in the tiny-data regime on MNIST/EMNIST and achieve near-human performance in Omniglot and QuickDraw only-one-shot learning.
Our model also enables $k$-means-style clustering to capture human-level visual abstractions in human-intuitive archetype generations.
This paper is a first step towards a general approach to a comprehensive, human-like framework for human-level performance in diverse applications.
We list a few future generalizations below.

Consider two general types of images:
1) images of abstract patterns, or \emph{abstract images}, \eg those of symbols and doodles;
2) photorealistic images of real-world objects, or \emph{real-world images}, \eg those in CIFAR10/100~\citep{KrizhevskyH2009}.
This paper focuses on the first type, handling abstract images by modeling humans' distortion-based intuition.
For real-world images, it may be more efficient to first model cognitive simplification and then apply our current distortion model.
Humans have remarkable visual abstraction ability to classify real-world images by first converting them into abstract icons or \emph{e (picture)+moji (character)}s (\eg the emoji of a face, the outline of a mountain, the shape of a lake) and then comparing these simplifications~\citep{SingerSKH2022}.
Following this, an efficient way to apply our method to real-world images is to follow this pipeline---preprocessing them first into ``emojis'' and then comparing ``emojis'' with distortion.
Stylized or abstract images, such as giraffe doodles and hieroglyphics, are those that can be treated as ``emojis'' already.
There are baselines to attempt first, \eg smart edge detectors~\citep{XieT2015}, but the human visual system does more than edge detection.
In future work, we aim for a complete theory of icon or ``emoji'' creation mimicking human capacity in order to deal with real-world images, 3D objects, and more.

Although our model has shown dominant classification performance in the tiny-data regime of the presented benchmarks, its dominance diminishes when training size increases.
This is due to $1$-NN being completely biased towards a single nearest neighbor and hence fragile against noisy, erroneous, and ambiguous training examples.
This suggests another future direction where our distortable canvas model may be designed jointly with a new, human-like classifier that introduces a small amount of learning into classification.
The goal is to achieve state-of-the-art results on all training sizes, which is not merely about swapping existing classifiers in and out.

\begin{figure}[t]
\begin{center}
\includegraphics[width=0.95\columnwidth]{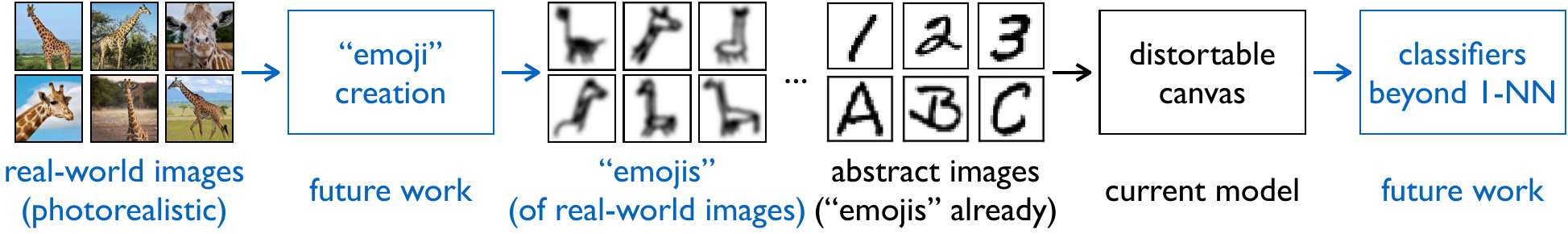}
\end{center}
\caption{Generalization of our distortable canvas model for two major future directions (two ends).}
\label{fig:research-pipeline}
\end{figure}

In order to create data- and energy-efficient ML prepared to face novel challenges that necessarily involve data-sparsity like growing children, artificial general intelligence (AGI) must replace experience with reason.
Inspired by this goal, AGI cannot simply rely on black-box models, but must maintain interpretability by thinking in the human way, where we learn functions from small training sets to drive cognitive simulations of similarity.
Figure~\ref{fig:research-pipeline} summarizes the pipeline for generalizing our current distortable canvas model to more general similarity simulations that reveal more insights about human-level visual abstractions.
These insights can in turn advance our understanding about various aspects of human cognition and facilitate learning and communication~\citep{FanBCW2023}.

\bibliography{abrv,conf_abrv,looos}
\bibliographystyle{iclr2022_conference}

\appendix

\section{Smooth Image on Distortable Canvas}
\label{sec:smooth-image-on-distortable-canvas}

We introduce a \emph{distortable canvas} model: any image is thought of as smoothly painted on an elastic (\eg rubber) canvas that can be bent, stretched, etc.
We further introduce \emph{canvas transformations} that can flexibly ``distort'' an image as we naturally simulate in our mind.
More specifically, we define a {\color{black} \emph{smooth image}} by a piecewise differentiable
$\mcal: \reals^2 \to \nonneg$, where $\reals^2$ denotes an infinite \emph{canvas} and $\nonneg$ denotes \emph{color} (grayscale in this paper).
We define a {\color{black}\emph{canvas transformation}} by $\alpha: \reals^2 \to \reals^2$, which ``reshapes'' the underlying canvas of a smooth image.
Examples include translation, rotation, scaling, and more.
We also define a {\color{black}\emph{color transformation}} by $\chi: \nonneg \to \nonneg$, which ``repaints'' a color.
We simplify color transformation and only use it to adjust image contrast via affine $\chi(c) := ac+b$.
In contrast, we do \emph{not} restrict canvas transformation, but consider \emph{all} 2D transformations.
Given $\mcal, \alpha, \chi$, the composition $\chi \circ \mcal \circ \alpha$ denotes the transformed image of $\mcal$ by transformations $\alpha,\chi$.

To mimic human intuition about general-appearance similarity, we introduce {\color{black} \emph{canvas distortion}}~$\dcal_V(\alpha)$ for any canvas transformation $\alpha$ and {\color{black} \emph{color distortion}} $\dcal_C(\mcal,\mcal')$ between two smooth images $\mcal,\mcal'$.
Our idea is to search for a transformation that mimics what humans naturally do to transform one image into another.
That is, a low-distorted $\alpha$ which makes little difference in color between $\mcal$ and the transformed $\mcal'$, or more precisely, an $\alpha$ that minimizes both $\dcal_V(\alpha)$ and $\dcal_C(\mcal,\ \chi \circ \mcal' \circ \alpha)$.

\textbf{Digital and smoothed images.}
An $m\times n$ \emph{digital image} is a discrete
$\msf: [m] \times [n] \to [0,1]$,
where $[k] := \{0,1,\ldots, k-1\}$ for any $k \in \integers$.
We call $[m] \times [n]$ the \emph{canvas grid} and any $\xg \in [m] \times [n]$ a \emph{grid point}.
For any $m\times n$ digital image $\msf$, we smooth it to $\mcal$ via a \emph{sum of kernels}:
\begin{align}
\label{eqn:smoothed-image}
\mcal(x) :=
\sum_{\xg \in [m] \times [n]} \msf(\xg)\cdot \kappa(\rho(\xg,x))
\quad \mbox{ for any } x \in \reals^2,
\end{align}
where a kernel $\kappa: \nonneg \to \nonneg$ is a decaying function (\eg linear, polynomial, Gaussian decay) and $\rho$ is a metric on $\reals^2$ (\eg $\ell_1, \ell_2, \ell_\infty$).
In this paper, we use linear decay and $\ell_\infty$, \ie $\kappa(\rho(\xg,x)) = 1-\frac{1}{\rho_c}\|z-x\|_\infty$ if $\|z-x\|_\infty < \rho_c$ (for some cutoff radius $\rho_c > 0$) and $\kappa(\rho(\xg,x)) = 0$ otherwise.
Note: $\mcal$ is defined everywhere on $\reals^2$.
This differs from Gaussian blurring as we do not discretize kernels.
It is key to use the smoothed image as input, which allows computing gradients analytically.
As such, we always smooth any digital image at first and then only manipulate the smoothed image.

\textbf{Arbitrary canvas transformations.}
We consider all 2D transformations (including those without a formula), but how do we represent them in a computer?
With respect to the \emph{standard grid} $[m] \times [n]$, we use the \emph{transformed grid} $\alpha([m] \times [n])$ to represent $\alpha$ digitally.
Thus, any canvas transformation $\alpha$ is \emph{digitally represented by} ($\deq$) a matrix $\alphamat \in \reals^{(mn)\times 2}$ whose $i$th row is the 2D coordinate of the transformed $i$th grid point.
We use the lexicographical order of a 2D grid, \eg with respect to $[2]\times [3]$, the identify transformation $\id \deq \idmat = [[0,0],[0,1],[0,2],[1,0],[1,1],[1,2]]$.
Any transformed image $\mcal \circ \alpha \deq \mcal(\alphamat) := (\ \mcal(\alphamat_0), \ \ldots, \ \mcal({\alphamat_{(mn-1)}})\ ) \in \reals^{(mn)}$, \ie a (vectorized) digital image sampled from $\mcal$ at the transformed grid $\alphamat$.

\begin{figure}[t]
\begin{center}
\includegraphics[width=0.75\columnwidth]{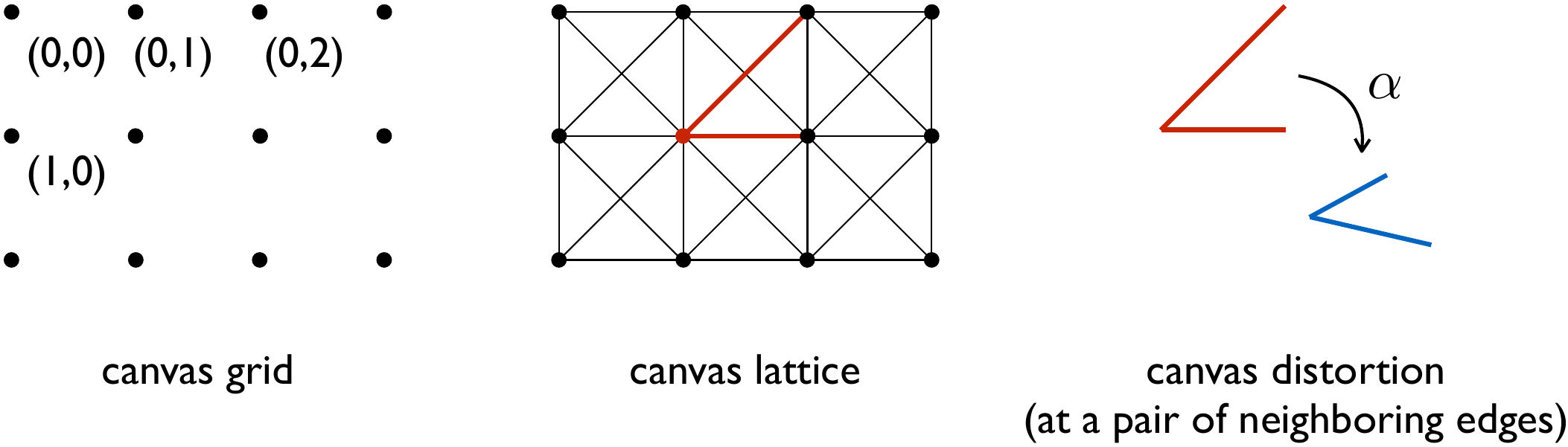}
\end{center}
\caption{Canvas grid and its corresponding lattice. Local distortions are computed at every pair of neighboring edges. One example of neighboring edges is highlighted in red.}
\label{fig:canvas-grid-lattice-distort}
\end{figure}

\textbf{Color and canvas distortions.}
The color distortion $\dcal_C$ measures the color discrepancy between $\mcal(\idmat)$ and $\mcal'(\alphamat)$ up to an affine color transformation $\chi$.
The canvas distortion $\dcal_V$ measures the distortion between the original grid $\idmat$ and the transformed grid $\alphamat$.
Formally,
\begin{align}
\label{eqn:color-distortion}
&{\color{black}\dcal_C(\mcal,\ \chi \circ \mcal' \circ \alpha) \deq \dcal_C(\ \mcal(\idmat),\ \chi(\mcal'(\alphamat))\ )} := \|a\mcal'(\alphamat)+b -\mcal(\idmat)\|_2^2 \\
\label{eqn:canvas-distortion}
&{\color{black}\dcal_V(\alpha) \deq \dcal_V(\idmat, \alphamat)} := \max_{\{\{i,j\},\{i',j'\}\} \in B_E}\left|\diffloglen^\alphamat_{\{i,j\}} - \diffloglen^\alphamat_{\{i',j'\}}\right|, \
\diffloglen^\alphamat_{\{i,j\}} := \log\dfrac{\|\alphamat_i-\alphamat_j\|_2}{\|\idmat_i-\idmat_j\|_2}
\end{align}
Here, $B_E$ comprises all pairs of neighboring edges in a \emph{canvas lattice} (introduced below).
Eq.~(\ref{eqn:canvas-distortion}) is derived from the mathematical definition of \emph{distortion of a function} by discretizing it across the canvas lattice.
This formula measures how far an arbitrary transformation is from being \emph{conformal}, which is flexible for \emph{local} isometries and scaling.
Given a canvas grid $[m]\times [n]$, its corresponding \emph{canvas lattice} is an undirected graph $L= (V,E)$, with the set of vertices $V = [m]\times [n]$ and the set of edges obtained by connecting neighboring vertices in the $\ell_\infty$ sense: $E = \{\{i,j\} \mid \|v_i - v_j\|_\infty = 1 \mbox{ for } v_i,v_j \in V\}$.
We say two edges are \emph{neighbors} if they form a $45\degree$ angle (Figure~\ref{fig:canvas-grid-lattice-distort}).

\textbf{General-appearance distance via distortion minimization.}
To minimize color and canvas distortions (\ref{eqn:color-distortion}) and (\ref{eqn:canvas-distortion}), we consider two dual views: minimizing $\dcal_C$ among low-distorted $\alpha$'s or minimizing $\dcal_V$ among best-matching $\alpha$'s.
We write the two views as the following two constrained optimization problems, together with their respective unconstrained equivalents: with $\epsilon \to 0_+$ and $\mu \to 0_+$,
\begin{align}
\label{eqn:dc-view}
\underset{\alpha,\chi}{\mbox{min.}}\ \ \dcal_C(\mcal,\ \chi \circ \mcal' \circ \alpha) \quad \mbox{ s.t. } \dcal_V(\alpha) \leq \epsilon \iff  \underset{\alpha,\chi}{\mbox{min.}}\ \ \dcal_V(\alpha) + \mu \dcal_C(\mcal,\ \chi \circ \mcal' \circ \alpha)\\
\label{eqn:dv-view}
\underset{\alpha,\chi}{\mbox{min.}}\ \ \dcal_V(\alpha) \quad \mbox{ s.t. } \dcal_C(\mcal,\ \chi \circ \mcal' \circ \alpha) \leq \epsilon \iff  \underset{\alpha,\chi}{\mbox{min.}}\ \ \dcal_C(\mcal,\ \chi \circ \mcal' \circ \alpha)+ \mu\dcal_V(\alpha)
\end{align}
We let the optima $\dcal_C^\star$ for (\ref{eqn:dc-view}) and $\dcal_V^\star$ for (\ref{eqn:dv-view}) denote two versions of our general-appearance distance that mimics human innate intuition.
We call them {\color{black}\emph{$\dcal_C$-distance}} and {\color{black}\emph{$\dcal_V$-distance}}, respectively.

\textbf{Transformation flow.}
To obtain both transformations and transformation processes that are human-like, we run (projected) gradient descent.
The iterative gradient steps yield not only a transformation $\alpha^\star$ in the end but also a {\color{black}\emph{transformation flow}} $\id = \alpha^{(0)} \to \alpha^{(1)} \to \cdots \to \alpha^{\star}$.
The resulting sequence of transformed images $\mcal' = \mcal' \circ \alpha^{(0)} \to \mcal' \circ \alpha^{(1)} \to \cdots \to \mcal'\circ \alpha^{\star} \approx \mcal$ (we omit $\chi$ for simplicity) makes up an animation (Figure~\ref{fig:transfs-and-flows}), which helps with human intuition on transforming $\mcal'$ to $\mcal$.
However, directly running (projected) gradient descent on (\ref{eqn:dc-view}) or (\ref{eqn:dv-view}) does not work, because it suffers from the curse of local minima, which we discuss and solve in the next section.

\section{Abstracted Multi-level Gradient Decent (AMGD)}
\label{sec:amgd}

The canvas distortion $\dcal_V$ is invariant under a variety of transformations (\eg $\dcal_V(\alpha) = 0$ for any conformal $\alpha$), which nicely mimics humans' flexible transformation options.
But this also implies lots of local/global minima and other critical points where the gradient is zero.
How much the color distortion $\dcal_C$ fluctuates as a function of $\alpha$ depends on the images $\mcal,\mcal'$.
But in most cases, $\dcal_C$ also has lots of local/global minima, the majority of which represent unwanted ``short cuts''---unnatural transformations that make $\dcal_C \to 0$ but would break the rubber canvas or create holes in it.
The curse of vanishing gradients can freeze gradient descent.
To unfreeze it, we lift gradient descent to higher levels, mimicking once again humans' abstraction power, as our internal optimization system is quite flexible in pursuing ``gradient-descent'' moves at multiple levels of abstraction.
We design two abstraction techniques:
a chain of anchor lattices to make hierarchical abstractions of canvas transformations and
a chain of color blurring to make hierarchical abstractions of image painting.

\textbf{Anchor grids and lattices.}
An {\color{black}\emph{anchor grid}} and its corresponding {\color{black}\emph{anchor lattice}} offer a simpler parameterization (\ie an abstraction) of canvas transformations.
Without such an abstraction, any transformed $[m] \times [n]$ grid $\alphamat \in \reals^{(mn)\times 2}$ consists of $2mn$ free parameters.
So, the optimization problems (\ref{eqn:dc-view}) and (\ref{eqn:dv-view}) are $2mn+2$ dimensional, which is not only computationally inefficient for large images but also has too much room for vanishing gradient.
We use a simpler $\alphamat$-parameterization that regularizes transformation, lowers distortion, and agrees with our intuition on rubber transformations.

\begin{figure}[t]
\begin{center}
\includegraphics[width=0.95\columnwidth]{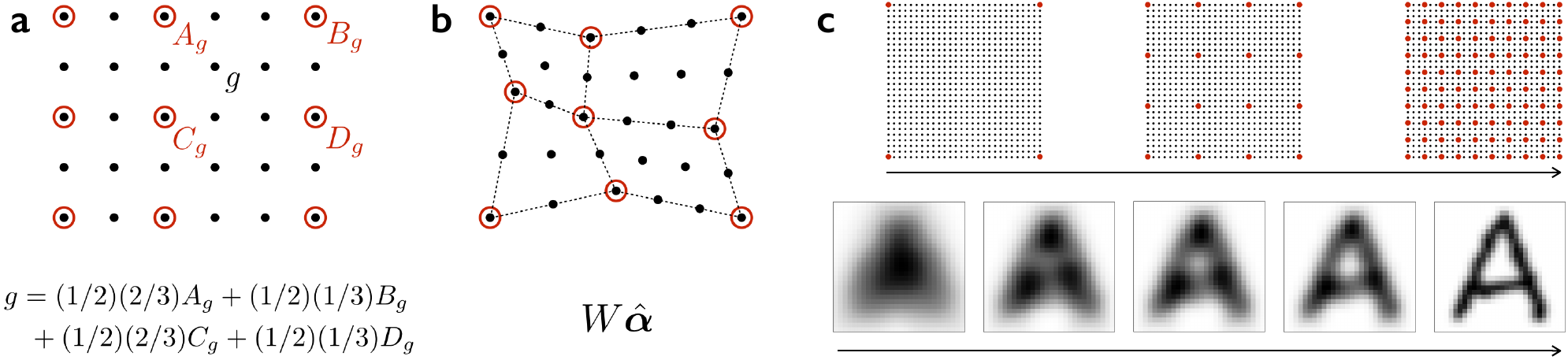}
\end{center}
\caption{An anchor system (a), its transformation (b), and a configuration of $(\hat{G}, \rho_c)$-solution path consisting of a chain of anchor grids/lattices and a chain of blurring (c).}
\label{fig:anchor-system}
\end{figure}

Formally, an \emph{anchor system} $(G,\hat{G}) = (M\times N,\hat{M} \times \hat{N})$ uses two layers of grids: an \emph{underlying grid} $G$ and an \emph{anchor grid} $\hat{G}$ atop, satisfying $\hat{M} \subseteq M, \hat{N}\subseteq N$, and $G \subseteq \cvxhull(\hat{G})$.
Figure~\ref{fig:anchor-system}a shows one example, where $G = [5]\times [6] = \{0, \ldots, 4\} \times \{0, \ldots, 5\}$ and $\hat{G} = \{0,2,4\}\times \{0,2,5\}$.
Under an anchor system, we can uniquely represent any grid point $g \in G$ via four anchors $A_g, B_g, C_g, D_g \in \hat{G}$ via proportional interpolation, or more precisely, the following double convex combination
\begin{align}
\label{eqn:double-conv-comb}
g = (1-\lambda_g)(1-\nu_g)A_g + (1-\lambda_g)\nu_g B_g + \lambda_g(1-\nu_g)C_g + \lambda_g\nu_g D_g.
\end{align}
Here, $A_gB_gD_gC_g$ can be uniquely selected as the smallest rectangle in ${\hat{G}}$'s lattice containing $g$; the two weight parameters $\lambda_g, \nu_g$ are computed based on relative position, \eg as in Figure~\ref{fig:anchor-system}a.
The relation between grid points and anchors can be summarized by a weight matrix $W \in \reals^{|G|\times |\hat{G}|}$.
Its $i$th row stores weights for the $i$th grid point (say $g$ in (\ref{eqn:double-conv-comb})) and contains at most four non-zero entries (\ie coefficients in (\ref{eqn:double-conv-comb})) located at the columns corresponding to $A_g,B_g,C_g,D_g$, respectively.

Given an anchor system $(G,\hat{G})$, any canvas transformation $\alpha \deq \alphamat \in \reals^{|G|\times 2}$ under $G$ and $\deq \hat{\alphamat} \in \reals^{|\hat{G}|\times 2}$ under $\hat{G}$.
$\hat{\alphamat}$ is a submatrix of $\alphamat$, which induces an equivalence relation on the set of all canvas transformations: $\alphamat, \betamat$ are equivalent iff $\hat{\alphamat} = \hat{\betamat}$, and $\hat{\alphamat}$ abstracts the equivalence class $\{\betamat \mid \hat{\betamat} = \hat{\alphamat}\}$.
Based on the maximum entropy principle~\citep{Jaynes1957}, a reasonable selection of a representative of this equivalence class is $W\hat{\alphamat}$, because $W\hat{\alphamat} \in \{\betamat \mid \hat{\betamat} = \hat{\alphamat}\}$ and evenly distributes the transformed grid points.
Figure~\ref{fig:anchor-system}b illustrates this type of even distribution, which agrees with human intuition on how a rubber surface would naturally react when transforming forces are applied at anchors.

Using an anchor system in optimization problems (\ref{eqn:dc-view}) and (\ref{eqn:dv-view}) adds very little to computing distortions and gradients: we reuse the computation with $\alphamat = W\hat{\alphamat}$ and perform only one additional chain-rule step ${\partial \alphamat}/{\partial \hat{\alphamat}} = W$.
By doing so, however, the number of optimization variables in (\ref{eqn:dc-view}) or (\ref{eqn:dv-view}) reduces from $|G|+2$ to $|\hat{G}|+2$ (\eg if $G = [28]\times [28]$ and $\hat{G} = \{0,27\}\times \{0,27\}$, the number reduces from $1570$ to $10$).
It is important to note that using a simpler anchor grid is \emph{not} the same as downsampling.
If it were, one would plug in $\alpha \leftarrow \hat{\alphamat}$, but we plug in $\alpha \leftarrow W\hat{\alphamat}$.
In our case, image colors are still sampled from the underlying grid rather than downsampled from the anchor grid.
So, using our anchor system is not information lossy while still benefiting from reduced optimization size.
Running gradient descent (w.r.t. anchors) in abstracted optimization spaces effectively bypasses critical points.

\textbf{Blurring.}
Another view to lifting gradient descent to a high-level, abstracted optimization space, is to blur the image.
Intuitively, blurring ignores low-level fluctuation, similar to how humans naturally abstract an image.
Blurring helps remedy vanishing gradients and is done in our image smoothing process.
The cutoff radius $\rho_c$ in $\kappa$ in (\ref{eqn:smoothed-image}) controls the blurring extent: larger $\rho_c$ means more blurred.

\textbf{Abstracted multi-level gradient descent.}
Mixing the two abstraction techniques yields our AMGD technique proceeding from higher- to lower-level abstractions.
Given an anchor grid $\hat{G}$ and a cutoff radius $\rho_c$, we denote the corresponding (\ref{eqn:dc-view}) and (\ref{eqn:dv-view}) by $DC(\hat{G},\rho_c)$ and $DV(\hat{G},\rho_c)$, respectively.
For either, we solve for a $(\hat{G},\rho_c)$-solution path, from coarser $\hat{G}$ and larger $\rho_c$ to finer $\hat{G}$ and smaller $\rho_c$.
Let $\hat{G}_k$ be a $k\times k$ evenly distributed anchor grid and $\hat{L}_k$ be its corresponding lattice.
Figure~\ref{fig:anchor-system}c shows a chain of anchor lattices $\{\hat{L}_{3^i+1}\}_{i=0,1,2,\ldots}$ and cutoff radii $\{\eta^j\rho_{c_0}\}_{j=0,1,2,\ldots}$.
It is easy initially to align two blurred blobs via small canvas adjustments, implying a small number of iterations to converge to $\dcal_C \approx \dcal_V \approx 0$.
As we proceed along the solution path, the images restore more detail but the finer $\hat{L}_k$ helps manage that detail.
In a solution path, an earlier solution is used to \emph{warm start} the subsequent solve step, which further alleviates the curse of vanishing gradients.
Notably, even the starting $\hat{L}_2$ comprising only four corner anchors parameterizes a large family of transformations containing all affine transformations.
Finer anchor grids/lattices express more flexible transformations (including local, global, piecewise affine, and more), approaching human-level flexibility.

\section{Transformation Multi-flows}
\label{sec:transformation-multi-flows}

To group $N$ smooth images $\mcal_1, \ldots, \mcal_N$ into $K$ clusters, we solve the optimization problem below:
\begin{align}
\label{eqn:dc-view-multi-flow}
\underset{\substack{\alpha_1,\ldots, \alpha_N \\ \overline{\alpha}_1, \ldots, \overline{\alpha}_K \\ C_1, \ldots, C_K}}{\mbox{minimize}}\ \ \sum_{k=1}^K\sum_{i \in C_k}\dcal_C(\overline{\mcal}_{k}\circ \overline{\alpha}_{k},\  \mcal_i \circ \alpha_i) \quad \mbox{ subject to } \sum_{i=1}^N\dcal_V(\alpha_i) \leq \epsilon,
\end{align}
where $C_k$ denotes the $k$th cluster, $\overline{\mcal}_k \circ \overline{\alpha}_k$ denotes the $k$th centroid, and $\mcal_i \circ \alpha_i$ denotes the $i$th transformed image flowing to its corresponding centroid together with all other $N-1$ transformed images.
One can check that (\ref{eqn:dc-view-multi-flow}) is an extension of (\ref{eqn:dc-view}) where we omitted $\chi$ for simplicity.
Solving (\ref{eqn:dc-view-multi-flow}) is similar to $k$-means via \emph{alternating refinement}:
\begin{itemize}[leftmargin=0.25in]
\item the assignment step assigns each transformed image $\mcal_i \circ \alpha_i$ to $C_{k^\star}$ according to
\begin{align*}
k^\star = \argmin_{k=1, \ldots, K}\dcal_C(\overline{\mcal}_{k}\circ \overline{\alpha}_{k},\  \mcal_i \circ \alpha_i);
\end{align*}
\item the update step solves (\ref{eqn:dc-view-multi-flow}) for one gradient-descent step given the $C_k$s.
\end{itemize}
Upon convergence, we obtain $C_1^\star, \ldots, C_K^\star$ as clusters and $\overline{\mcal}_1 \circ \overline{\alpha}_1, \ldots, \overline{\mcal}_K \circ \overline{\alpha}_K$ as centroids.

\end{document}